\documentclass[pmlr,twocolumn,10pt]{jmlr} 

\usepackage{booktabs}
\usepackage{siunitx}

\theorembodyfont{\upshape}
\theoremheaderfont{\scshape}
\theorempostheader{:}
\theoremsep{\newline}

\jmlrvolume{LEAVE UNSET}
\jmlryear{2022}
\jmlrsubmitted{LEAVE UNSET}
\jmlrpublished{LEAVE UNSET}
\jmlrworkshop{Conference on Health, Inference, and Learning (CHIL) 2022} 

\title[Impact of class imbalance on chest x-ray classifiers]{Impact of class imbalance on chest x-ray classifiers: \\towards better evaluation practices for discrimination and calibration performance}
\author{%
\Name{Candelaria Mosquera}
\Email{candelaria.mosquera@hospitalitaliano.org.ar}\\
\addr Universidad Tecnológica Nacional, Argentina. Hospital Italiano de Buenos Aires, Argentina.
\AND
\Name{Luciana Ferrer}
\Email{lferrer@dc.uba.ar}\\
\addr Universidad de Buenos Aires, Argentina
\AND
\Name{Diego Milone} \Email{dmilone@sinc.unl.edu.ar}\\
\addr Universidad Nacional del Litoral, Argentina
\AND
\Name{Daniel Luna} \Email{daniel.luna@hospitalitaliano.org.ar}\\
\addr Hospital Italiano de Buenos Aires, Argentina
\AND
\Name{Enzo Ferrante} \Email{eferrante@sinc.unl.edu.ar}\\
\addr Universidad Nacional del Litoral, Argentina
}

\begin{document}

\maketitle

\begin{abstract}
This work aims to analyze standard evaluation practices adopted by the research community when assessing chest x-ray classifiers, particularly focusing on the impact of class imbalance in such appraisals. Our analysis considers a comprehensive definition of model performance, covering not only \textit{discriminative performance} but also \textit{model calibration}, a topic of research which has received increasing attention during the last years within the machine learning community. Firstly, we conducted a literature study to analyze common scientific practices and confirmed that: (1) even when dealing with highly imbalanced datasets, the community tends to use metrics that are dominated by the majority class; and (2) it is still uncommon to include calibration studies for chest x-ray classifiers, albeit its importance in the context of healthcare. Secondly, we perform a systematic experiment on two major chest x-ray datasets to explore the behavior of several performance metrics under different class ratios and show that widely adopted metrics can conceal the performance in the minority class. Finally, we recommend the inclusion of complementary metrics to better reflect the performance of the system in such scenarios. Our study indicates that current evaluation practices adopted by the research community for chest x-ray computer-aided diagnosis systems may not reflect their performance in real clinical scenarios, and suggest alternatives to improve this situation.
\end{abstract}

\paragraph*{Data and Code Availability}
This paper uses the ChestX-ray14 dataset \citep{nihdataset} from the National Institutes of Health, and the CheXpert \citep{chexpert} dataset from Stanford University. Both databases are publicly available; the links to access them are available in the cited references. Our source code is fully available online at \url{https://github.com/code-anonym/imbalanceCXR}.


\section{Introduction}
\label{sec:intro}
The application of machine learning to medical images has grown rapidly in the last years, showing high levels of performance \citep{bigData,AIhealth,bigDataHealth}. As the advancements in medical AI translate into real-world clinical tools, the capabilities of these systems are being reconsidered, since traditional methods for evaluating performance seem insufficient to describe their impact on the diagnosis pathway \citep{beede,beam2020challenges}. Measuring and reporting performance in a useful way is a complex task, influenced by the characteristics of the dataset and the intended use of the algorithm.

Class imbalance is a distinctive characteristic of medical datasets, where the rare events that are important to detect have significantly fewer observations than the normal or healthy cases. In this domain, the performance of classifiers should be assessed with metrics that adequately reflect the performance on both classes, that is, those metrics not dominated by the system's behavior on the majority class \citep{luque2019imbalanceConfusionMatrix,garcia2012imbalancePreprocess,orriols2009imbalanceRuleBased}.

\begin{figure*}[t]
\centering
\includegraphics[width=0.8\textwidth]{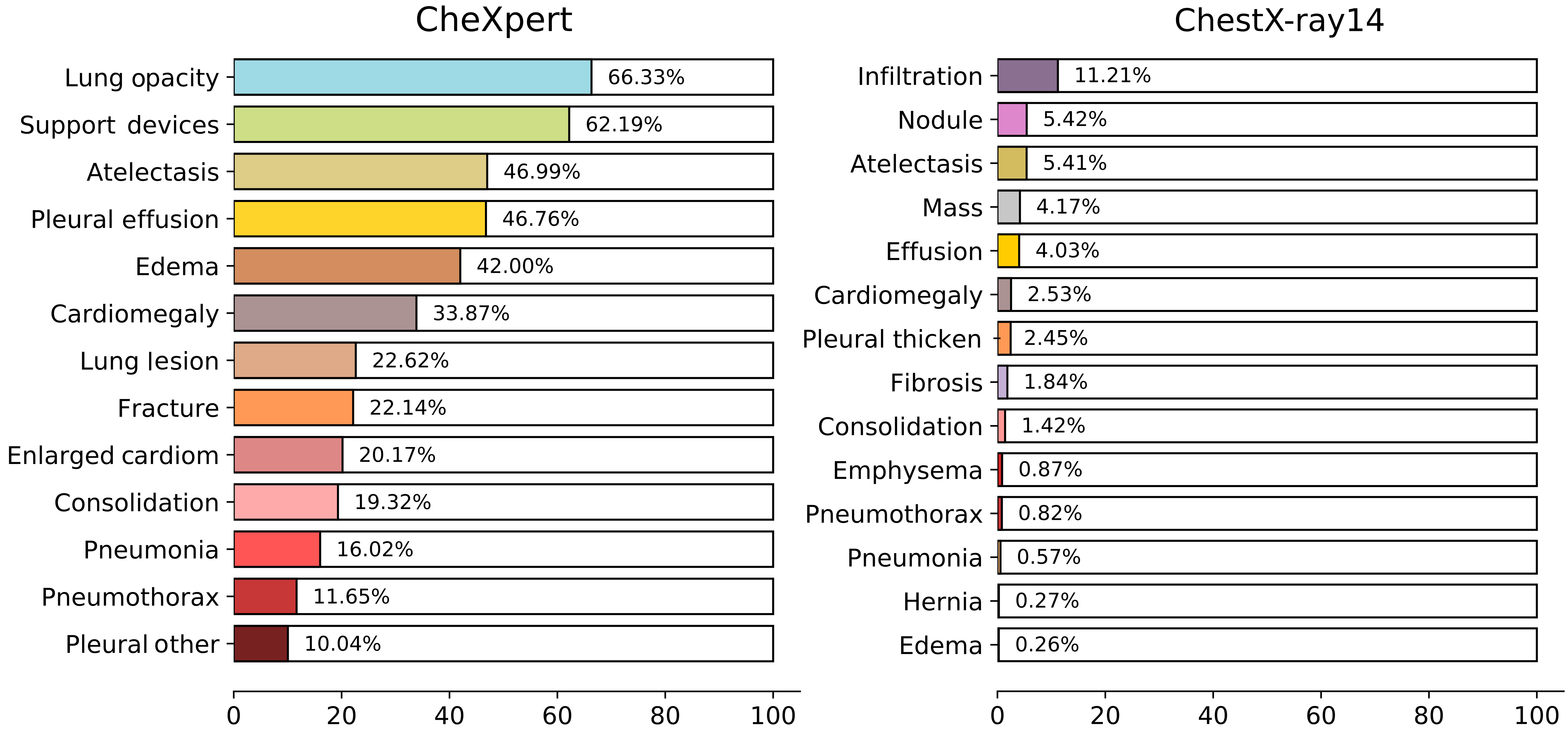}
  
  \caption{Positive class ratio for the radiological findings annotated in two major chest x-ray datasets. The colored area represents the proportion of positive observations for each finding. These values were computed using only one image per patient and only frontal views. Uncertain and empty labels were excluded.}
    \label{fig:classes}
\end{figure*}
The analysis of chest x-ray images is a typical case of an imbalanced classification problem. The low prevalence of radiological findings is evidenced in the large datasets that were publicly released lately to encourage the development of deep learning (DL) algorithms for chest x-ray diagnosis \citep{ccalli2021deep} (see examples in Figure \ref{fig:classes}). These datasets contain between 100,000 and 380,000 images with their accompanying diagnosis labels. In particular, ChestX-ray14 \citep{nihdataset} and CheXpert \citep{chexpert} are well-known and widely adopted datasets, containing class labels automatically assigned with Natural Language Processing methods (NLP) from radiological reports. In this work, we are interested in analysing the implications of the high class imbalance in chest x-ray datasets when evaluating image classifiers, both in terms of discrimination and calibration.

Extensive work has been published on the use of public chest x-ray datasets to train DL classifiers, and many studies report good discrimination metrics \citep{ccalli2021deep}. These studies use the area under the Receiver Operating Characteristics curve (AUC-ROC) as the primary outcome, and conclusions are based mainly on this metric. The AUC-ROC has been adopted as a standard metric in the academic community of medical image computing because it presents several practical advantages for the evaluation of binary classifiers. As this metric does not depend on the class ratios of the test dataset, it simplifies the comparison of models evaluated in datasets with different prevalence. Moreover, the ROC curve provides a comprehensive evaluation of the discrimination performance of a model over all possible operating points \citep{erickson2021magician}. As we will discuss in this paper and as previous studies suggest in other application scenarios, ROC curves require special caution when used for imbalanced datasets \citep{lopez2013imbalanceClassification,PLOS_roc, ozenne2015precision,sahiner2017comparison,learningImbalance,davis2006relationship}. 

The assessment of chest x-ray classifiers could be harmed by the adoption of AUC-ROC as the central performance metric, considering the highly imbalanced nature of the problem. In this work, we provide empirical evidence of this adoption by conducting a literature analysis of articles on chest x-rays automated classification, published during 2019 and 2020. We then provide empirical evidence of possible limitations of AUC-ROC on imbalanced scenarios through an experimental study on CheXpert and ChestX-ray14, and compare it with the area under the precision-recall curve (AUC-PR), showing that this metric better reflects the performance of the model on both the majority and the minority class.

Another distinctive characteristic of the medical imaging domain is the need for interpretable outputs to assist in clinical decision-making. In this regard, quantifying model uncertainty becomes crucial \citep{uncertaintyNature}. A system is said to be well-calibrated when its output reflects the uncertainty about a sample's class given the input information \citep{ovadia2019can,dawid1982well}; for example, when the output is given by the probability that the sample belongs to the class of interest given the input \citep{focalLosscalib}. Model calibration is well studied in epidemiology literature \citep{van2016calibration,collins2019calibration}, and although it is also a topic of active research in the machine learning community, our literature analysis shows that it is not commonly discussed in medical imaging works, particularly in the problem of chest x-ray diagnosis. 

Classification performance should be understood as the combination of discrimination and calibration capabilities \citep{blattenberger1985separating}. Popular evaluation methods such as the ROC plot and the confusion matrix are measurements of discrimination performance only. While monotonic transformations of the output of the system will result in no changes in the ROC plots, they do affect the calibration of such output. Metrics that are minimized when the algorithm reports the posterior probability of the class (given the input features, the modeling assumptions and the available training data) are known as proper scoring rules \citep{gneiting2007strictly}. These metrics, such as the negative log likelihood or the Brier score  \citep{brier1950verification}, are affected both by the discrimination and the calibration performance of the model. However, the impact of class imbalance should also be considered when evaluating calibration. If imbalance is large, traditional proper scoring rules might show good values because they are dominated by the majority class, but they can hide undesirable behavior in the minority class. We discuss the use of an alternative indicator, the Balanced Brier score, and show that this metric can help to detect performance issues in low-prevalence pathologies.

This study addresses the task of performance assessment in imbalanced data for chest x-ray image classification systems in two steps: (1) a literature analysis that characterizes the common practices adopted by the scientific community when reporting results; (2) a systematic experiment with a DL classifier trained and tested on reference datasets, to compare various performance metrics.

\section{Literature Analysis}
For our literature analysis, we included conference and journal research articles from two sources:
\begin{enumerate}
    \item \textbf{PubMed}: we selected articles containing the expression “chest radiograph” or “chest x-ray” and the expressions “artificial intelligence”, “machine learning” or “deep learning” in the title, published in 2019 or 2020. 
    \item \textbf{Proceedings from the International Conference on Medical Image Computing and Computer Assisted Intervention (MICCAI)}: we selected articles that used chest x-rays as test set from the volumes “X-ray imaging” and “Computer-aided diagnosis” from the 2019 edition and the volumes “Machine Learning methodologies”, “Prediction and diagnosis”, “Machine Learning applications” and “Heart and lung imaging” from the 2020 edition. 
\end{enumerate}

\begin{figure*}[t!]
  \includegraphics[width=0.99\linewidth]{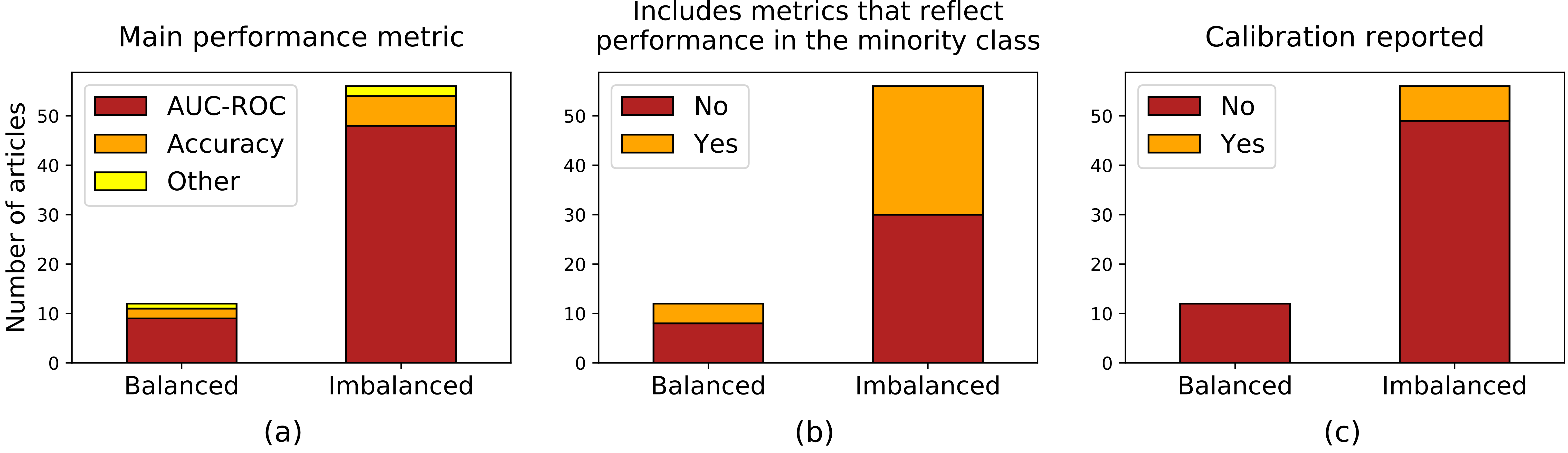}

  \caption{Results from the literature analysis on chest x-ray classification. Comparison between articles with balanced and imbalanced test sets for three aspects of performance assessment.}
  \label{fig:literature}
  
\end{figure*}
We excluded reviews, editorial notes, author corrections, erratas or commentary publications, articles that do not inform the imbalance ratio of the test set or do not report classification results on chest x-ray images. We also excluded articles focusing on COVID-19 because it has been recently reported that many papers published during 2020 applying machine learning to detect and prognosticate for COVID-19 using chest radiographs present poor-quality data, poor application of machine learning methodology, poor reproducibility and introduce biases in study design \citep{roberts2021common}. 
We revised the articles and checked the following aspects:
\begin{itemize}
    \item The class ratios on the test set. If the minority class represents less than 33\% of images (imbalance ratio of 1:3), it was categorized as “Imbalanced"; otherwise as "Balanced". 
    \item The main metric for reporting classification performance. This was determined by the following conditions: the primary outcome of the study as mentioned by the authors, else the first metric mentioned in the abstract, or the metric with most detailed reporting in figures or tables. 
    \item Whether the study reports at least one of the following metrics, which more adequately reflect the performance of the system for the minority class: AUC-PR, positive predictive value (precision), negative predictive value, Matthews Correlation Coefficient or F-score. 
    \item Whether model uncertainty or calibration were numerically reported or discussed in some way.
\end{itemize}

\subsection{Results for the literature analysis}
PubMed and MICCAI searches resulted in 64 and 37 articles respectively. After exclusion, we evaluated 68 articles in total. A detailed table on this analysis is available as Supplementary Material. The primary study outcome is AUC-ROC in 57 works (83\%), accuracy in 8 (12\%), and F1 score, Brier score and recall in one article each. Moreover, 90\% of studies do not report performance metrics that are affected by calibration, indicating that this aspect was not assessed. 
Regarding class ratios, 79\% of articles use imbalanced test sets, suggesting this is a common scenario in chest x-rays analysis. Figure \ref{fig:literature} compares performance evaluation aspects between articles with balanced and imbalanced test sets. 96\% of studies with an imbalanced test set (50 of 54) use AUC-ROC or accuracy as main metric (Figure 2a), and 56\% (30 of 54) report only performance metric that are dominated by the majority class (Figure 2b), failing to assess model behaviour in the minority class. Model calibration is considered in 17\% of these works (9 of 54), while no studies with balance test sets discuss it (Figure 2c).

Our analysis confirms our two initial hypothesis:
\begin{enumerate}
    \item That even when dealing with highly imbalanced datasets, the community tends to focus on AUC-ROC and metrics dominated by the majority class; and
    \item It is still uncommon to include calibration studies for CAD systems, albeit its importance in the context of healthcare.

\end{enumerate}
\section{Experimental study}
Once we established the standard evaluation practises adopted by the research community through our literature analysis, we proceeded to perform a systematic experimental study to assess the behavior of various performance metrics across pathologies with different class ratios. Data processing and model training were performed with Pytorch\footnote{Torch 1.7.0 (https://pytorch.org/). Torchvision 0.8.1 (https://pytorch.org/vision/).}, adapting an open framework by \citep{cohen2020limits}. Our source code is fully available online.\footnote{https://github.com/code-anonym/imbalanceCXR}

\subsection{Data}
We used two well-known databases, the ChestX-ray14 \citep{nihdataset}, from the National Institutes of Health, and the CheXpert \citep{chexpert} from Stanford University. Both datasets were automatically annotated applying Natural Language Processing tools on radiological reports. ChestX-ray14 is labeled for the presence of 14 findings as positive or negative, while CheXpert dataset is labeled for the presence of 13 findings as positive, negative, or uncertain. The two datasets have six classes in common. We used frontal chest x-rays and included only one image per patient. In the case of CheXpert we excluded images with positive “Support Devices” label. The class imbalance of these two datasets was measured separately for each annotated radiological finding, considering the presence of pathology as the positive class (Figure \ref{fig:classes}).

As preprocessing we cropped non-squared images to a squared central crop, resized to 224x224 pixels, and scaled pixel values to [-1024,1024], as is standard for these datasets \citep{Rajpurkar2017chestClassif}. We performed data augmentation on the training set, according to the best results reported in \citep{cohen2019chester}: random rotation of up to 45°, combined with scaling and translation of up to 15\% of image size. 20\% of images from each dataset were selected as the test set, and the remaining set was further split into 20\% for tuning (validation) and 80\% for training. The experiment was repeated five times for each dataset, using five different random seeds for the splitting. 

\subsection{Model}
We used a DenseNet-121 architecture\citep{densenet}, as it has been used by previous systematic analysis in chest x-rays \citep{Rajpurkar2017chestClassif,cohen2020limits,larrazabal2020gender}. We used a final dense layer of 11 and 14 sigmoid units for CheXpert and ChestX-Ray14, respectively. Sigmoid activation sets up the model for a binary classification task for each class. We followed a standard training configuration as reported by \citep{cohen2020limits}. Convolutional layers weights were initialized with Kaiming normal distribution. We trained the model for 100 epochs with batch size of 64, using Adam optimizer (weight decay of 1e-5) and binary cross entropy loss (with no class weight). In the CheXpert dataset, cases with an uncertain or empty label were not considered for loss calculation. The initial learning rate was 1e-3 and it was reduced by a factor of 10 every 40 epochs. 
\begin{figure*}[t!]
  \includegraphics[width=0.99\linewidth]{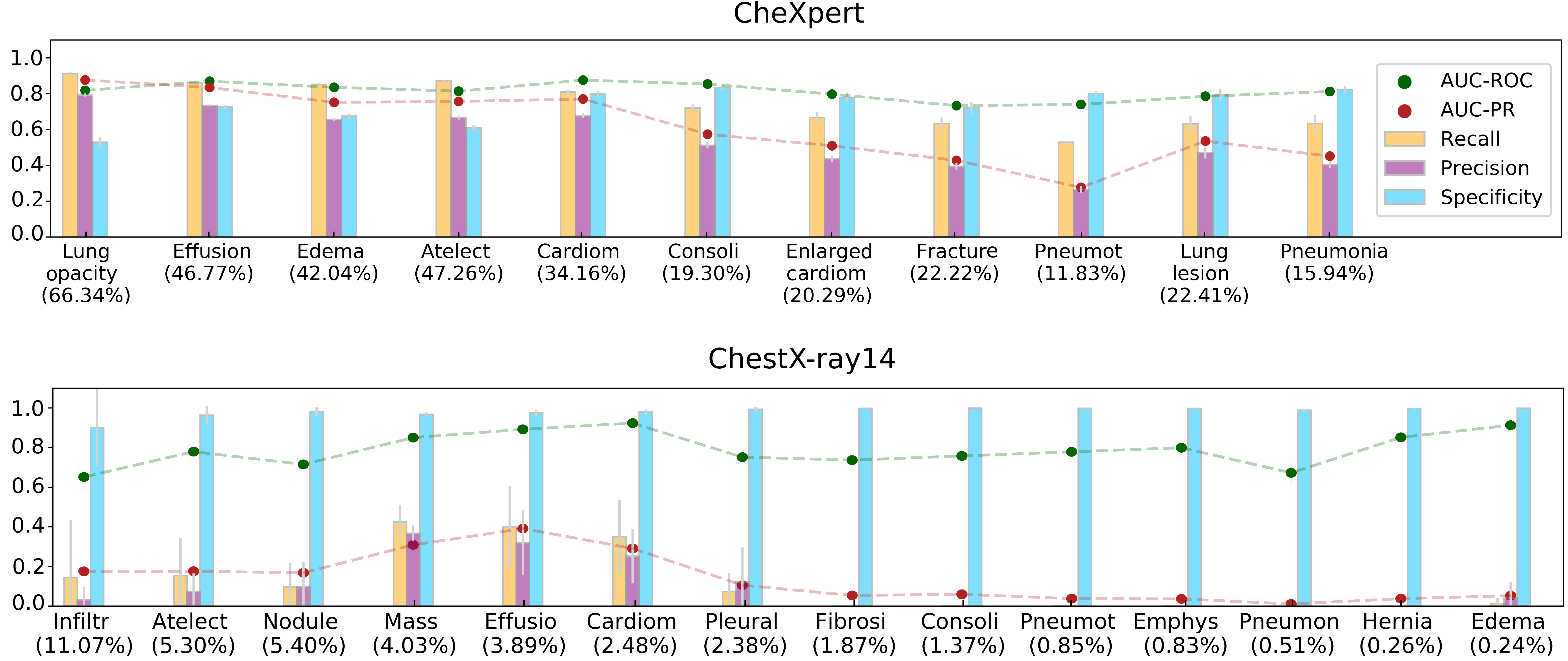}

  \caption{Discrimination metrics as imbalance increases. Values in parentheses indicate the average positive class ratio in the test set.}
  \label{fig:discrimination}
  
\end{figure*}
\subsection{Performance metrics}
We computed the average ROC and PR curves for the test set of both datasets across 5 runs. The area under the curves was calculated using the trapezoidal rule. To calculate metrics that require a threshold (recall, specificity and precision) we chose the operating point for each pathology as the threshold that maximized F1 score in the tuning set. Performance metrics were calculated with the Scikit Learn package\footnote{Scikit Learn 0.23.2 (https://scikit-learn.org/)}.
We calculated the Brier score \citep{brier1950verification} for each binary classification task (i.e., each pathology class). This metric is defined as the average squared difference between the true class and the estimated class probability of each image in the test set: 
$$Brier = \frac {1}{N}\sum_{i=1}^{N}(y_i-\hat{y}_i)^2$$
where $y_i$ is the binary class label for the $i^{th}$ image (0 or 1) and  $y_i$ is its output on the corresponding sigmoid unit of the model.

As Brier is a proper scoring rule, its optimal value corresponds to a perfect prediction: a system that is perfectly calibrated and perfectly discriminative will have a Brier score of zero. However, the Brier score could be misleading in imbalanced datasets: performance may be good on average over all samples, but poor for the minority class. A dummy model that outputs a zero probability score for all samples would have a Brier score of 0.01 in a test set with one positive sample and 99 negatives. It would look like a good system, though it is virtually useless. To address this problem, \citet{brierpositivo} propose a decomposition into stratified Brier scores, calculating separately the Brier score over the positive samples (Brier$^+$) and over the negative samples (Brier$^-$) 
$$Brier^{+} = \frac{\sum_{y_i=1}(y_i-\hat{y}_i)^2}{N_{pos}}$$
$$Brier^{-} = \frac{\sum_{y_i=0}(y_i-\hat{y}_i)^2}{N_{neg}}$$
Here, an interesting alternative is the sum of both stratified Brier scores, which can be used for evaluation or as objective function for optimization algorithms: 
$$BalancedBrier = Brier^{+} +Brier^{-}$$

\subsection{Experimental results}
We present results measured in the held-out test set as an average of five runs of training for each dataset, using different random seeds for splitting sets and following standard training configurations for a DenseNet model. Figure \ref{fig:discrimination} shows discrimination metrics for all pathologies, sorted by decreasing positive class ratio. Pathologies with similar high values of AUC-ROC show a large difference in AUC-PR when the class imbalance is very different (for example, atelectasis and pneumonia in CheXpert, or effusion and hernia in ChestX-ray14). This suggests that evaluating the AUC-PR curve is especially important to understand model behavior in imbalanced datasets. If only the AUC-ROC was considered, we would conclude that the model is equally good for most tasks, while the actual behavior is significantly different. For example, the AUC-ROC values for edema and effusion on the ChestX-ray14 dataset are close (0.91 and 0.89, respectively), showing the discrimination performance for these pathologies is similar. However, their AUC-PR values differ greatly (0.05 vs. 0.39), suggesting the system has a poor usability for edema detection. A high AUC-ROC value might not imply an acceptable performance in highly imbalanced classification tasks.
\begin{figure*}[t]
  \includegraphics[width=0.99\linewidth]{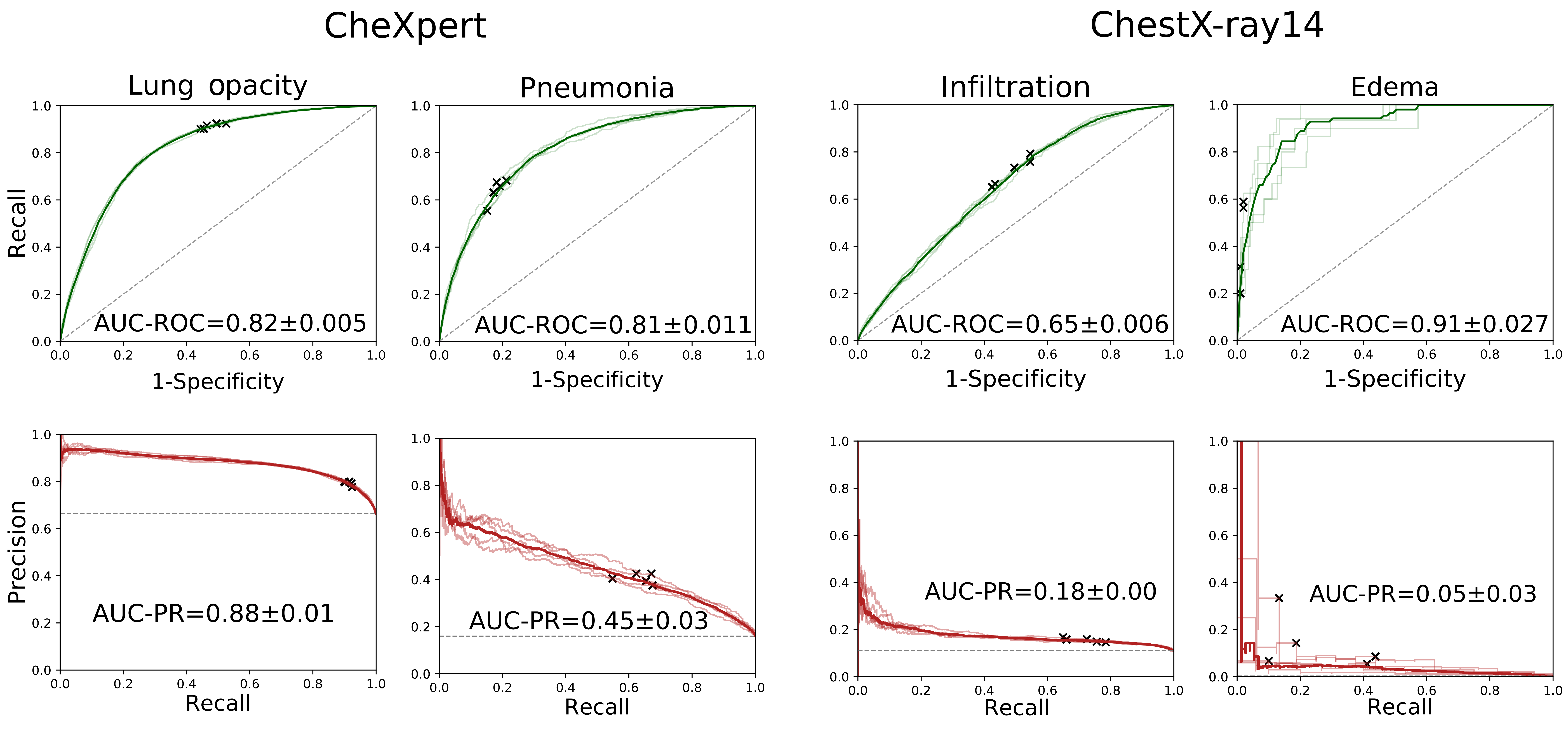}

  \caption{ ROC curve (top) and PR curve (bottom) for the pathologies with highest and lowest imbalance ratio. Black crosses indicate the operating point corresponding to maximum F1 score for each run. Bold line represents the mean across runs.}
  \label{fig:curves}
  
\end{figure*}

We observe that specificity is high across pathologies: in spite of a large number of false positives, the much larger number of true negatives dominates the numerator of its formula. On the other hand, precision tends to decrease along with the positive class ratio: as precision does not consider true negatives, it is greatly affected by false positives. The same specificity value might not imply a comparable performance in imbalanced datasets; instead, precision becomes more informative in these scenarios. 
\begin{figure*}[t]
  \includegraphics[width=0.99\linewidth]{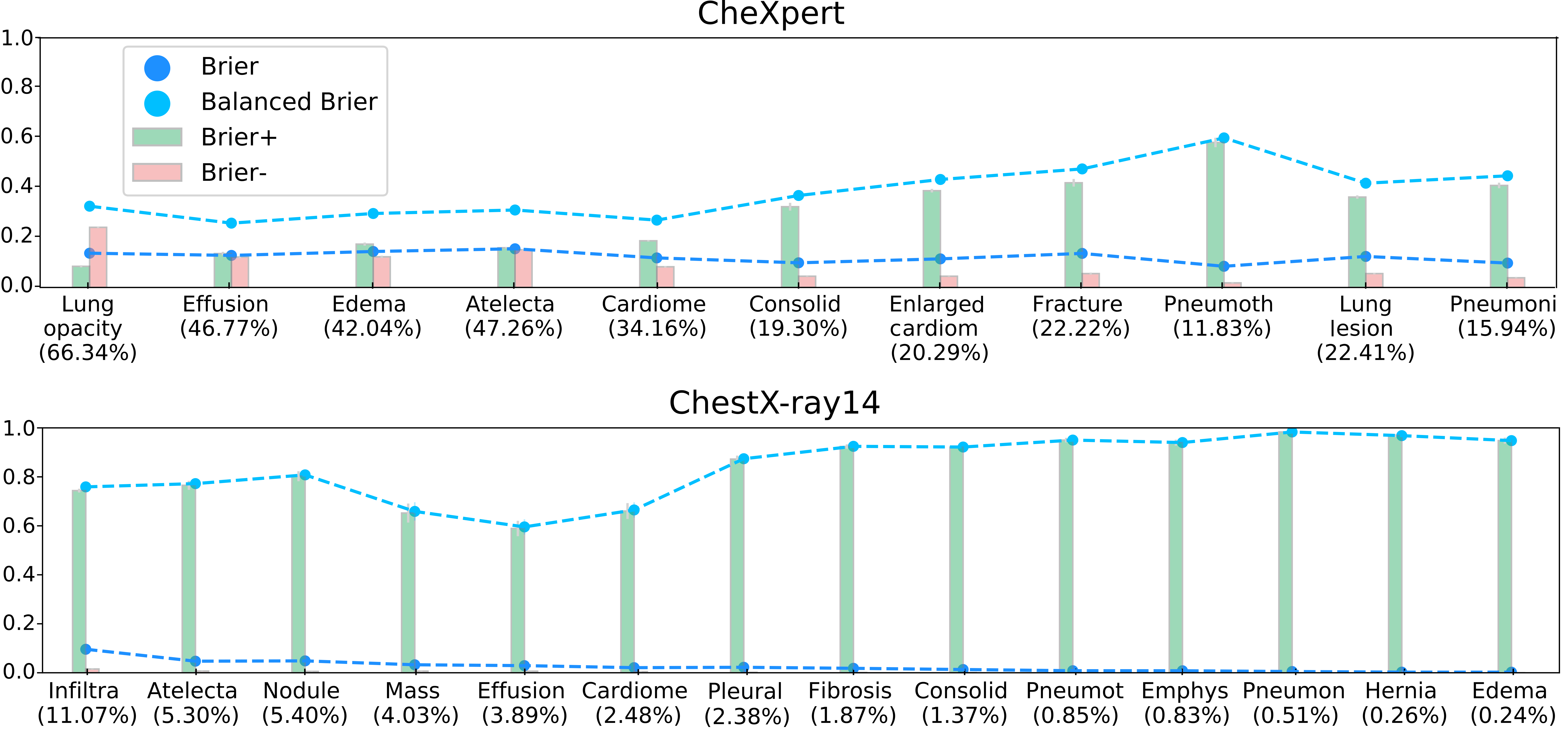}

  \caption{Brier, Balanced Brier and stratified Brier scores in test, as positive class ratio decreases. Values in parentheses indicate the average positive class ratio for each class.}
  \label{fig:briers}
  
\end{figure*}

Figure \ref{fig:curves} shows ROC and PR curves for the pathologies with highest and lowest positive class ratio in each dataset. We can see how lung opacity and pneumonia detection have similar ROC plots, but their PR plots are strikingly different. Regarding ChestX-Ray14, the detection of edema presents a high mean AUC-ROC value, however the AUC-PR is very low. The operating points with maximum F1 score in the validation sets are illustrated as black crosses on curves. For strongly imbalanced classes, there is a greater variation of performance across split seeds, showing that the dependence on specific positive samples is stronger as imbalance increases. 

Given fixed score distributions for each class, decreasing the positive class ratio does not change the AUC-ROC but it does affect the AUC-PR: for the same recall level the precision will get worse. The AUC-ROC of a random classifier would be 0.5 regardless of the positive class ratio in the test set, while its AUC-PR would vary along with this ratio \citep{PLOS_roc}. In imbalanced scenarios, a significant portion of the ROC curve could correspond to operating points with a large number of false positives (i.e., points with low precision). As a consequence, the AUC-ROC is highly dominated by the performance on operating points that would not be acceptable in an actual application, making it an inadequate metric for such scenarios. On the other hand, in the PR curve, those points with a large number of false positives correspond to low values of precision and a lower height of the curve, reducing the value of the AUC-PR.

Figure \ref{fig:briers} shows the average Brier scores for each class. Unlike AUC metrics, which measure only discrimination power, the Brier score is affected also by model calibration, representing a comprehensive assessment of performance. However, if only the standard Brier score was considered, we would conclude that the model has a similar  performance across pathologies. Instead,  we see that Brier$^+$ is higher than Brier$^-$ for pathologies with low positive class ratio. This shows that performance is consistently worse for the minority class than the majority one across pathologies. A central finding of this work is showing that a problem in the classification of the minority class might only be noted when using the balanced version of Brier. Using the standard Brier score, this performance problem would go unnoticed. The Balanced Brier score can be used to measure overall performance in imbalanced data: a high score implies that the quality of the class probabilities generated by the system will not yield useful bayesian decisions, which could be caused by poor calibration, poor discrimination, or a combination of both—either in the minority or the majority class, or both.

\section{Discussion}

The central objective of this work was to study how performance is affected by the positive class ratios of chest pathologies in the context of chest x-ray classification. We systematically trained a convolutional neural network with two large chest x-ray datasets and evaluated test performance by comparing AUC-ROC and AUC-PR, as well as Brier and Balanced Brier. We observed that AUC-ROC alone can be misleading: the fact that the ROC curve includes multiple operating points with low precision values makes it less informative for imbalanced scenarios than the PR curve. This experimental study shows the importance of comparing various performance metrics during model assessment, as the AUC-ROC alone can lead to false conclusions.

The comparison of AUC-ROC and AUC-PR has been studied in several prior works \citep{luque2019imbalanceConfusionMatrix,garcia2012imbalancePreprocess,orriols2009imbalanceRuleBased}. \citet{ozenne2015precision} showed that AUC-PR had a higher correlation with precision across simulations with various imbalance ratios, and concluded that it reflected better the discriminant ability of a biomarker in rare diseases. \citet{sahiner2017comparison} compared the statistical power of these two metrics, and indicated that AUC-PR can offer a statistical advantage when the positive class ratio is low, while in a balanced scenario AUC-ROC has slightly higher power. The fact that the PR curve is prevalence-dependent has practical implications: it hinders the comparison across datasets with different class ratios, and requires a test set with a prevalence that matches that of the true population to make valid statistical inferences. Although both issues can be solved with an adequate experimental design — as they can be addressed by reporting the test class ratios and by applying properly-designed correction methods, respectively — AUC-ROC allows a more straight-forward analysis. Another appealing characteristic of the ROC and PR plots is their visual interpretability, as they provide an overview of performance across a wide range of operating points. However, \citet{PLOS_roc} showed that, in the context of imbalanced datasets, ROC plots can be deceptive, due to an intuitive but mistaken interpretation of specificity. Therefore, AUC-PR should be reported alongside AUC-ROC to provide a comprehensive assessment of the discrimination power of a classifier. Although these aspects might be well-known in the machine learning community, our literature analysis reveals that it is not usually addressed in x-ray image classification studies. The literature search was not exhaustive; however, we consider that the selected sample of articles is representative of the current common practices in the field of chest x-ray automated diagnosis.

Both AUC-ROC and AUC-PR are ranking metrics, as they depend on the probability that a randomly chosen positive instance is ranked higher than a randomly chosen negative instance. For this reason, they do not consider the problem of threshold optimization. This can be a limitation in clinical practice, where obtaining a final binary classification is critical to assist in decision-making. The selection of an optimal operating point is a key aspect of model performance \citep{brierpositivo}. Metrics that are calculated for a fixed operating point—such as the confusion matrix—can grade the quality of the decisions that are based on this threshold. They should be reported to describe the model’s actual utility for a classification task, and class imbalance should be considered when interpreting them. Recall and specificity are metrics that are not affected by the imbalance ratio: when the negative class predominates strongly over the positive class, high values of these metrics can be hiding a great number of false positives. On the contrary, precision reveals this problem when it occurs. Although this has been previously reported \citep{erickson2021magician,PLOS_roc} and it was further confirmed in our experimental results, we observed that several works still fail to report precision (positive predictive value) when evaluating chest x-rays classifiers. 

Although the calibration aspect is critical for the successful translation of DL models to clinical practice \citep{uncertaintyNature}, our literature review showed that most articles do not address this issue. Proper scoring rules can be used to consider this aspect, as they quantify performance including both discrimination and calibration. These metrics measure the quality of the class probabilities generated by the system for each sample assuming they will be used to make optimal decisions using Bayes decision theory \citep{ramos2018deconstructing}. Standard proper scoring rules, such as the Brier score, are affected by large class imbalance and become dominated by the performance in the majority class, as was confirmed by our experimental study. To address this problem, the Balanced Brier score is a useful metric in imbalanced datasets, which reflects performance in both the minority and the majority class.

\section{Future works and conclusions}
Future work includes extending the experimental study to other large chest x-ray datasets, such as MIMIC and Padchest, and use other neural architectures. Moreover, training strategies that address class imbalance, such as weighted loss functions or enrichment sampling techniques,  should be applied to explore their effect on performance for the minority class. To assess the calibration performance of a model, independently of its discrimination performance, we should compare its overall performance against that of an ideally calibrated version of the model’s outputs for the test set (i.e., applying a monothonic transformation that conserves the same discrimination power), obtained for example with the PAV algorithm \citep{ramos2018deconstructing}. Moreover, the behavior with varying positive class ratios across sets should be studied, as a difference between train and test ratios would need specific techniques to adapt the model operating point in a cost-effective way.

The main conclusion of this study is that a comprehensive assessment of a classifier in imbalanced scenarios should include metrics not dominated by the majority class, which should be consistently reported alongside AUC-ROC, such as the AUC-PR; and incorporate metrics affected by calibration, such as the Balanced Brier score suggested in this work. We hope that this work will contribute to improve evaluation practices for CAD systems by improving the comprehension of model behaviour in real scenarios, and ultimately help to translate medical AI into the clinical setting.

\section*{Institutional Review Board (IRB)}
This research does not require IRB approval.


\bibliography{jmlr-sample}

\begin{thebibliography}{35}
\providecommand{\natexlab}[1]{#1}
\providecommand{\url}[1]{\texttt{#1}}
\expandafter\ifx\csname urlstyle\endcsname\relax
  \providecommand{\doi}[1]{doi: #1}\else
  \providecommand{\doi}{doi: \begingroup \urlstyle{rm}\Url}\fi

\bibitem[Beam and Kohane(2018)]{bigDataHealth}
Andrew~L Beam and Isaac~S Kohane.
\newblock Big data and machine learning in health care.
\newblock \emph{Jama}, 319\penalty0 (13):\penalty0 1317--1318, 2018.

\bibitem[Beam et~al.(2020)Beam, Manrai, and Ghassemi]{beam2020challenges}
Andrew~L Beam, Arjun~K Manrai, and Marzyeh Ghassemi.
\newblock Challenges to the reproducibility of machine learning models in
  health care.
\newblock \emph{Jama}, 323\penalty0 (4):\penalty0 305--306, 2020.

\bibitem[Beede et~al.(2020)Beede, Baylor, Hersch, Iurchenko, Wilcox,
  Ruamviboonsuk, and Vardoulakis]{beede}
Emma Beede, Elizabeth Baylor, Fred Hersch, Anna Iurchenko, Lauren Wilcox,
  Paisan Ruamviboonsuk, and Laura~M. Vardoulakis.
\newblock A human-centered evaluation of a deep learning system deployed in
  clinics for the detection of diabetic retinopathy.
\newblock In \emph{Proceedings of the 2020 CHI Conference on Human Factors in
  Computing Systems}, CHI '20, page 1–12, New York, NY, USA, 2020.
  Association for Computing Machinery.
\newblock \doi{10.1145/3313831.3376718}.

\bibitem[Blattenberger and Lad(1985)]{blattenberger1985separating}
Gail Blattenberger and Frank Lad.
\newblock Separating the brier score into calibration and refinement
  components: A graphical exposition.
\newblock \emph{The American Statistician}, 39\penalty0 (1):\penalty0 26--32,
  1985.

\bibitem[Brier(1950)]{brier1950verification}
Glenn~W Brier.
\newblock Verification of forecasts expressed in terms of probability.
\newblock \emph{Monthly weather review}, 78\penalty0 (1):\penalty0 1--3, 1950.

\bibitem[{\c{C}}all{\i} et~al.(2021){\c{C}}all{\i}, Sogancioglu, van Ginneken,
  van Leeuwen, and Murphy]{ccalli2021deep}
Erdi {\c{C}}all{\i}, Ecem Sogancioglu, Bram van Ginneken, Kicky~G van Leeuwen,
  and Keelin Murphy.
\newblock Deep learning for chest x-ray analysis: A survey.
\newblock \emph{Medical Image Analysis}, page 102125, 2021.

\bibitem[Cohen et~al.(2019)Cohen, Bertin, and Frappier]{cohen2019chester}
Joseph~Paul Cohen, Paul Bertin, and Vincent Frappier.
\newblock Chester: A web delivered locally computed chest x-ray disease
  prediction system.
\newblock \emph{arXiv preprint arXiv:1901.11210}, 2019.

\bibitem[Cohen et~al.(2020)Cohen, Hashir, Brooks, and
  Bertrand]{cohen2020limits}
Joseph~Paul Cohen, Mohammad Hashir, Rupert Brooks, and Hadrien Bertrand.
\newblock On the limits of cross-domain generalization in automated x-ray
  prediction.
\newblock In \emph{Medical Imaging with Deep Learning}, 2020.
\newblock URL \url{https://arxiv.org/abs/2002.02497}.

\bibitem[Collins and Moons(2019)]{collins2019calibration}
Gary~S Collins and Karel~GM Moons.
\newblock Reporting of artificial intelligence prediction models.
\newblock \emph{The Lancet}, 393\penalty0 (10181):\penalty0 1577--1579, 2019.

\bibitem[Davis and Goadrich(2006)]{davis2006relationship}
Jesse Davis and Mark Goadrich.
\newblock The relationship between precision-recall and roc curves.
\newblock In \emph{Proceedings of the 23rd international conference on Machine
  learning}, pages 233--240, 2006.

\bibitem[Dawid(1982)]{dawid1982well}
A~Philip Dawid.
\newblock The well-calibrated bayesian.
\newblock \emph{Journal of the American Statistical Association}, 77\penalty0
  (379):\penalty0 605--610, 1982.

\bibitem[Erickson and Kitamura(2021)]{erickson2021magician}
Bradley~J Erickson and Felipe Kitamura.
\newblock Magician’s corner: 9. performance metrics for machine learning
  models, 2021.

\bibitem[Garc{\'\i}a et~al.(2012)Garc{\'\i}a, S{\'a}nchez, and
  Mollineda]{garcia2012imbalancePreprocess}
Vicente Garc{\'\i}a, Jos{\'e}~Salvador S{\'a}nchez, and Ram{\'o}n~Alberto
  Mollineda.
\newblock On the effectiveness of preprocessing methods when dealing with
  different levels of class imbalance.
\newblock \emph{Knowledge-Based Systems}, 25\penalty0 (1):\penalty0 13--21,
  2012.

\bibitem[Gneiting and Raftery(2007)]{gneiting2007strictly}
Tilmann Gneiting and Adrian~E Raftery.
\newblock Strictly proper scoring rules, prediction, and estimation.
\newblock \emph{Journal of the American statistical Association}, 102\penalty0
  (477):\penalty0 359--378, 2007.

\bibitem[Huang et~al.(2017)Huang, Liu, Van Der~Maaten, and
  Weinberger]{densenet}
Gao Huang, Zhuang Liu, Laurens Van Der~Maaten, and Kilian~Q Weinberger.
\newblock Densely connected convolutional networks.
\newblock In \emph{Proceedings of the IEEE conference on computer vision and
  pattern recognition}, pages 4700--4708, 2017.

\bibitem[Irvin et~al.(2019)Irvin, Rajpurkar, Ko, Yu, Ciurea-Ilcus, Chute,
  Marklund, Haghgoo, Ball, Shpanskaya, et~al.]{chexpert}
Jeremy Irvin, Pranav Rajpurkar, Michael Ko, Yifan Yu, Silviana Ciurea-Ilcus,
  Chris Chute, Henrik Marklund, Behzad Haghgoo, Robyn Ball, Katie Shpanskaya,
  et~al.
\newblock Chexpert: A large chest radiograph dataset with uncertainty labels
  and expert comparison.
\newblock In \emph{Proceedings of the AAAI Conference on Artificial
  Intelligence}, volume~33, pages 590--597, 2019.

\bibitem[Kompa et~al.(2021)Kompa, Snoek, and Beam]{uncertaintyNature}
Benjamin Kompa, Jasper Snoek, and Andrew~L Beam.
\newblock Second opinion needed: communicating uncertainty in medical machine
  learning.
\newblock \emph{NPJ Digital Medicine}, 4\penalty0 (1):\penalty0 1--6, 2021.

\bibitem[Krawczyk(2016)]{learningImbalance}
Bartosz Krawczyk.
\newblock Learning from imbalanced data: open challenges and future directions.
\newblock \emph{Progress in Artificial Intelligence}, 5\penalty0 (4):\penalty0
  221--232, 2016.

\bibitem[Larrazabal et~al.(2020)Larrazabal, Nieto, Peterson, Milone, and
  Ferrante]{larrazabal2020gender}
Agostina~J Larrazabal, Nicol{\'a}s Nieto, Victoria Peterson, Diego~H Milone,
  and Enzo Ferrante.
\newblock Gender imbalance in medical imaging datasets produces biased
  classifiers for computer-aided diagnosis.
\newblock \emph{Proceedings of the National Academy of Sciences}, 117\penalty0
  (23):\penalty0 12592--12594, 2020.

\bibitem[L{\'o}pez and Fern{\'a}ndez(2013)]{lopez2013imbalanceClassification}
V~L{\'o}pez and A~Fern{\'a}ndez.
\newblock A, garcia, s., palade, v. \& herrera, f.(2013). an insight into
  classification with imbalanced data: Empirical results and current trends on
  using data intrinsic characteristics.
\newblock \emph{Information Sciences}, 250\penalty0 (11):\penalty0 113--141,
  2013.

\bibitem[Luque et~al.(2019)Luque, Carrasco, Mart{\'\i}n, and de~las
  Heras]{luque2019imbalanceConfusionMatrix}
Amalia Luque, Alejandro Carrasco, Alejandro Mart{\'\i}n, and Ana de~las Heras.
\newblock The impact of class imbalance in classification performance metrics
  based on the binary confusion matrix.
\newblock \emph{Pattern Recognition}, 91:\penalty0 216--231, 2019.

\bibitem[Mukhoti et~al.(2020)Mukhoti, Kulharia, Sanyal, Golodetz, Torr, and
  Dokania]{focalLosscalib}
Jishnu Mukhoti, Viveka Kulharia, Amartya Sanyal, Stuart Golodetz, Philip~HS
  Torr, and Puneet~K Dokania.
\newblock Calibrating deep neural networks using focal loss.
\newblock \emph{arXiv preprint arXiv:2002.09437}, 2020.

\bibitem[Orriols-Puig and
  Bernad{\'o}-Mansilla(2009)]{orriols2009imbalanceRuleBased}
Albert Orriols-Puig and Ester Bernad{\'o}-Mansilla.
\newblock Evolutionary rule-based systems for imbalanced data sets.
\newblock \emph{Soft Computing}, 13\penalty0 (3):\penalty0 213--225, 2009.

\bibitem[Ovadia et~al.(2019)Ovadia, Fertig, Ren, Nado, Sculley, Nowozin,
  Dillon, Lakshminarayanan, and Snoek]{ovadia2019can}
Yaniv Ovadia, Emily Fertig, Jie Ren, Zachary Nado, David Sculley, Sebastian
  Nowozin, Joshua~V Dillon, Balaji Lakshminarayanan, and Jasper Snoek.
\newblock Can you trust your model's uncertainty? evaluating predictive
  uncertainty under dataset shift.
\newblock \emph{arXiv preprint arXiv:1906.02530}, 2019.

\bibitem[Ozenne et~al.(2015)Ozenne, Subtil, and
  Maucort-Boulch]{ozenne2015precision}
Brice Ozenne, Fabien Subtil, and Delphine Maucort-Boulch.
\newblock The precision--recall curve overcame the optimism of the receiver
  operating characteristic curve in rare diseases.
\newblock \emph{Journal of clinical epidemiology}, 68\penalty0 (8):\penalty0
  855--859, 2015.

\bibitem[Raghupathi and Raghupathi(2014)]{bigData}
Wullianallur Raghupathi and Viju Raghupathi.
\newblock Big data analytics in healthcare: promise and potential.
\newblock \emph{Health information science and systems}, 2\penalty0
  (1):\penalty0 1--10, 2014.

\bibitem[Rajpurkar et~al.(2017)Rajpurkar, Irvin, Zhu, Yang, Mehta, Duan, Ding,
  Bagul, Langlotz, Shpanskaya, et~al.]{Rajpurkar2017chestClassif}
Pranav Rajpurkar, Jeremy Irvin, Kaylie Zhu, Brandon Yang, Hershel Mehta, Tony
  Duan, Daisy Ding, Aarti Bagul, Curtis Langlotz, Katie Shpanskaya, et~al.
\newblock Chexnet: Radiologist-level pneumonia detection on chest x-rays with
  deep learning.
\newblock \emph{arXiv preprint arXiv:1711.05225}, 2017.

\bibitem[Ramos et~al.(2018)Ramos, Franco-Pedroso, Lozano-Diez, and
  Gonzalez-Rodriguez]{ramos2018deconstructing}
Daniel Ramos, Javier Franco-Pedroso, Alicia Lozano-Diez, and Joaquin
  Gonzalez-Rodriguez.
\newblock Deconstructing cross-entropy for probabilistic binary classifiers.
\newblock \emph{Entropy}, 20\penalty0 (3):\penalty0 208, 2018.

\bibitem[Roberts et~al.(2021)Roberts, Driggs, Thorpe, Gilbey, Yeung, Ursprung,
  Aviles-Rivero, Etmann, McCague, Beer, et~al.]{roberts2021common}
Michael Roberts, Derek Driggs, Matthew Thorpe, Julian Gilbey, Michael Yeung,
  Stephan Ursprung, Angelica~I Aviles-Rivero, Christian Etmann, Cathal McCague,
  Lucian Beer, et~al.
\newblock Common pitfalls and recommendations for using machine learning to
  detect and prognosticate for covid-19 using chest radiographs and ct scans.
\newblock \emph{Nature Machine Intelligence}, 3\penalty0 (3):\penalty0
  199--217, 2021.

\bibitem[Sahiner et~al.(2017)Sahiner, Chen, Pezeshk, and
  Petrick]{sahiner2017comparison}
Berkman Sahiner, Weijie Chen, Aria Pezeshk, and Nicholas Petrick.
\newblock Comparison of two classifiers when the data sets are imbalanced: the
  power of the area under the precision-recall curve as the figure of merit
  versus the area under the roc curve.
\newblock In \emph{Medical Imaging 2017: Image Perception, Observer
  Performance, and Technology Assessment}, volume 10136, page 101360G.
  International Society for Optics and Photonics, 2017.

\bibitem[Saito and Rehmsmeier(2015)]{PLOS_roc}
Takaya Saito and Marc Rehmsmeier.
\newblock The precision-recall plot is more informative than the roc plot when
  evaluating binary classifiers on imbalanced datasets.
\newblock \emph{PloS one}, 10\penalty0 (3):\penalty0 e0118432, 2015.

\bibitem[Van~Calster et~al.(2016)Van~Calster, Nieboer, Vergouwe, De~Cock,
  Pencina, and Steyerberg]{van2016calibration}
Ben Van~Calster, Daan Nieboer, Yvonne Vergouwe, Bavo De~Cock, Michael~J
  Pencina, and Ewout~W Steyerberg.
\newblock A calibration hierarchy for risk models was defined: from utopia to
  empirical data.
\newblock \emph{Journal of clinical epidemiology}, 74:\penalty0 167--176, 2016.

\bibitem[Wallace and Dahabreh(2014)]{brierpositivo}
Byron~C Wallace and Issa~J Dahabreh.
\newblock Improving class probability estimates for imbalanced data.
\newblock \emph{Knowledge and information systems}, 41\penalty0 (1):\penalty0
  33--52, 2014.

\bibitem[Wang et~al.(2017)Wang, Peng, Lu, Lu, Bagheri, and Summers]{nihdataset}
Xiaosong Wang, Yifan Peng, Le~Lu, Zhiyong Lu, Mohammadhadi Bagheri, and
  Ronald~M Summers.
\newblock Chestx-ray8: Hospital-scale chest x-ray database and benchmarks on
  weakly-supervised classification and localization of common thorax diseases.
\newblock In \emph{Proceedings of the IEEE conference on computer vision and
  pattern recognition}, pages 2097--2106, 2017.

\bibitem[Yu et~al.(2018)Yu, Beam, and Kohane]{AIhealth}
Kun-Hsing Yu, Andrew~L Beam, and Isaac~S Kohane.
\newblock Artificial intelligence in healthcare.
\newblock \emph{Nature biomedical engineering}, 2\penalty0 (10):\penalty0
  719--731, 2018.

\end{thebibliography}






\end{document}